%% file: root.tex
\documentclass[a4paper, conference]{IEEEtran}  

\usepackage{xcolor}
\usepackage{booktabs}
\usepackage{graphicx} 
\usepackage{epsfig} 
\usepackage{mathptmx} 
\usepackage{times} 
\usepackage{amsmath} 
\usepackage{amssymb}  
\usepackage{tabularx}
\usepackage{hyperref}
\usepackage[caption=false]{subfig}
\usepackage{cite}
\usepackage{multirow}

\newcommand{\norm}[1]{\left\lVert#1\right\rVert}

\title{\LARGE \bf
Semi-automatic 3D Object Keypoint Annotation and Detection for the Masses
}
\author{\IEEEauthorblockN{Kenneth Blomqvist\IEEEauthorrefmark{1},
Jen Jen Chung\IEEEauthorrefmark{1},
Lionel Ott\IEEEauthorrefmark{1} and
Roland Siegwart\IEEEauthorrefmark{1}}
\IEEEauthorblockA{\IEEEauthorrefmark{1} Autonomous Systems Lab, ETH Z\"urich}
}

\begin{document}

\maketitle
\thispagestyle{empty}
\pagestyle{empty}

\begin{abstract}
Creating computer vision datasets requires careful planning and lots of time and effort. In robotics research, we often have to use standardized objects, such as the YCB object set, for tasks such as object tracking, pose estimation, grasping and manipulation, as there are datasets and pre-learned methods available for these objects. This limits the impact of our research since learning-based computer vision methods can only be used in scenarios that are supported by existing datasets.
In this work, we present a full object keypoint tracking toolkit, encompassing the entire process from data collection, labeling, model learning and evaluation. We present a semi-automatic way of collecting and labeling datasets using a wrist mounted camera on a standard robotic arm. Using our toolkit and method, we are able to obtain a working 3D object keypoint detector and go through the whole process of data collection, annotation and learning in just a couple hours of active time.
\end{abstract}

\section{INTRODUCTION}
\input{01-introduction}

\section{RELATED WORK}
\input{02-related-work}

\section{METHOD}
\input{03-method}

\section{EXPERIMENTS}
\input{04-experiments}

\section{RESULTS}
\input{05-results}

\section{DISCUSSION AND CONCLUSIONS}
\input{06-conclusions}

\bibliographystyle{IEEEtran}
\bibliography{references}

\end{document}

%% file: 01-introduction.tex
Most modern computer vision methods use large datasets to learn to predict features at run time. These have been demonstrated to enable many new capabilities in robotic object manipulation. While the methods are impressive, they are data hungry and require sizeable datasets of ground truth annotations to train. If we could quickly and cheaply create datasets, we could expand to more environments and enable many downstream tasks.

The data requirements force researchers of downstream robotics tasks to either use standard objects, for which trained models and computer vision pipelines have been made available, or a large investment has to be made upfront to collect and label a dataset. Creating a dataset requires either hand-labeling thousands of frames one-by-one, having a data collection setup with environment markers, as done in \cite{liu2020keypose}, or a tool such as LabelFusion \cite{marion2018label} can be used to partially automate the annotation process. However, LabelFusion requires mesh models of the objects. Creating a known model for objects in turn requires a high-fidelity object scanning setup, which is often unavailable. It also requires the objects to be rigid, or additional parameters need to be estimated to model deformation. Additionally, the objects and environment have to be such that depth sensors are able to accurately measure depth, excluding reflective or transparent objects.

In this paper, our goal is to track category-level semantic points in an object's coordinate frame relative to the camera frame for downstream robotic manipulation tasks. ``Category-level" meaning that objects vary, but the intra-category semantic meaning of keypoints are the same. Specifically, we want a system with the following properties:
\begin{enumerate}
    \item Can estimate 3D object keypoints on arbitrary objects
    \item Requires little effort to handle novel objects
    \item Can be used in the wild without having to use markers, motion tracking systems or otherwise modify the environment
    \item Does not rely on accurate depth sensing
    \item Can track multiple objects simultaneously in the image frame
\end{enumerate}

\begin{figure*}[ht]
    \centering
    \includegraphics[width=0.8\linewidth]{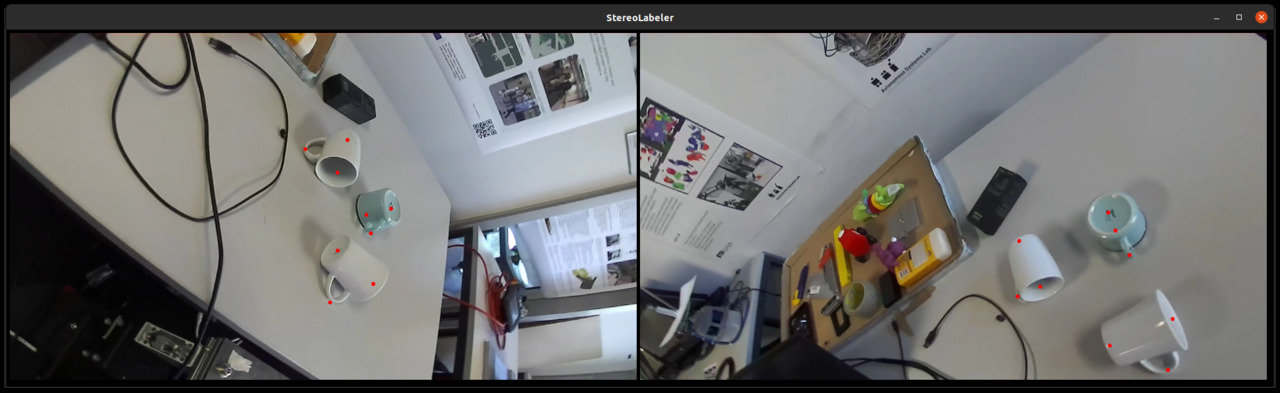}
    \caption{StereoLabel, our keypoint labeling tool. The user is presented with two images of the scene to label. The images are selected to maximize the orthogonality of the views.}\vspace{-0.5cm}
    \label{fig:stereo-label}
\end{figure*}

Existing methods, such as PVN3D \cite{pvn3d} and kPAM \cite{gao2019kpam} require semantic segmentation maps to train or they rely on external object instance segmentation. Semantic segmentation maps are time consuming to annotate, making the systems more expensive to deploy in new scenarios and for new objects.

In contrast, we present a complete 3D object keypoint tracking system, including both a learning-based object keypoint algorithm and a method to very quickly obtain the training labels needed by the algorithm.
Our method builds on the insight that we can forgo using semantic segmentation maps to distinguish between objects, if we instead introduce a center keypoint and predict a center map that associates each keypoint with a center keypoint. The amount of objects in the scene is inferred from the amount of detected center keypoints. This makes the labeling task a lot faster, as we can simply label 2D keypoints instead of having to also create dense instance segmentation masks.

We present a way to speed up data collection by capturing many views of the scene and propagating labels from two labeled viewpoints to all the others. We show that by calibrating our robot and making use of calibration and the kinematics of a robot arm, we can forgo using a motion tracking system or environment markers, as done by previous works. Using our system, data can be collected in the wild wherever our robot goes. This means that our tools can be deployed directly on the hardware intended for the downstream robotic task, streamlining the full problem definition and solution by avoiding additional steps. Calibrated robots are now commonly available and by using one, the data collection can be further automated and enables collecting data autonomously. 

We show two different versions of our learning-based algorithm that leverages our data collection pipeline to track keypoints of multiple objects in a scene. The first one uses both views of a stereo camera. The other one is a variation of the algorithm that can work with a monocular RGB camera.

We validate our method and tracking pipeline in experiments on two different object keypoint tracking scenarios. The first one is a single object valve tracking scenario. The second is a multiple object cup tracking task, showcasing that we can handle multiple objects simultaneously in a frame. We show that using only 22.5 minutes of recorded data across 45 sequences, and using less than 15 minutes of labeling time, we can learn a model that can track keypoints on objects of interest. We demonstrate that the resulting tracker is accurate enough to enable manipulation tasks, such as rotating a valve.

Code for our project is made available at \href{https://github.com/ethz-asl/object\_keypoints}{\texttt{github.com/ethz-asl/object\_keypoints}}.

%% file: 02-related-work.tex
\subsection{Datasets and Labeling Tools}

Several object pose datasets have emerged which use ground truth meshes. The most commonly used meshes are of the YCB object set \cite{calli2015ycb}. The YCB-video dataset from~\cite{posecnn} provides labeled 6D poses for objects in RGB images. The authors demonstrated that the dataset was capable of training their PoseCNN 6D object pose estimator. An initial estimate of the object poses from PoseCNN were used to generate the YCB-M dataset~\cite{grenzdorffer2020ycb}. This dataset was collected with seven depth cameras and they used fiducial markers and depth refinement to obtain the common frame of reference between cameras. While a robot arm was used to facilitate data collection, the authors did not use the kinematics of the robot nor did they use hand-eye calibration in the labeling process. Moreover, both of these datasets are limited to the YCB object set. \cite{tremblay2018deep} uses object models, simulation and rendering to obtain a dataset of ground truth object poses. 

While other labeling tools exist to create datasets of a priori unknown objects, these often have other limitations. \cite{stumpf2021salt} provides a semi-automated tool for creating 2D and 3D bounding box labels for multi-object scenes in RGB-D video. Their algorithm uses a GrabCut-based approach \cite{rother2004grabcut} to interpolate annotations over timesteps. However, the user still needs to adjust the propagated bounding boxes in each of the following frames. LabelFusion~\cite{marion2018label} can handle cluttered scenes, however, object meshes are required and must either be given or created manually using a scanning routine (e.g. with a handheld scanner or turntable). Our proposed method avoids this requirement altogether.

Finally, several methods train keypoint detectors using only a small set of labeled data. Simon et al.~\cite{simon2017hand} bootstrap a keypoint detection dataset for hand pose estimation using multiple views of the scene. The authors ensure that each iteration introduces new information via multiview geometry. However, because of this, performance is tied to the number of cameras in the setup. Multiview geometry is used by~\cite{yao2019monet} for human and animal pose estimation by deriving a differentiable semi-supervised loss function which is equivalent to minimizing epipolar divergence. They show that they can train a keypoint detection network using a large set of unlabeled images and comparatively few labeled images.

\subsection{Object Keypoints and 6D Pose Estimation}

Methods exist which predict keypoints, in 2D or 3D, to calculate the 6D object pose. Some estimate 2D points on an RGB image and solve for the pose using a PnP algorithm \cite{tekin2018, pix2pose, zakharov2019dpod, pvnet, Yinli2020}. Others predict keypoints directly in 3D space \cite{suwajanakorn2018discovery, pvn3d}.
6-PACK \cite{wang20206} presents a way to track single objects in real-time using keypoints which emerge in an unsupervised way. As the keypoints are learned end-to-end, additional components such as an attention mechanism are required in their keypoint tracking pipeline.
S3K \cite{vecerik2020s3k} is a self-supervised approach to learn semantic 3D keypoints. Similar to our approach, the authors use multiple camera views to propagate labels across images. However, in their case, they require a four-camera setup while our method is designed to work with a single camera.
NOCS \cite{wang2019normalized} uses a representation shared within an object category. The authors learn a model to regress to this representation from RGB and depth maps. 
PVN3D \cite{pvn3d} learns a model which produces semantic segmentation maps as well as per pixel keypoint and center votes from RGB-D frames. Similarly to PVN3D, we use a center prediction map to track multiple objects. However, PVN3D uses ground truth semantic instance segmentation maps to distinguish objects from each other, which are hard and expensive to label. We avoid this by associating keypoints directly with their corresponding object's center. This also circumvents the need for the expensive clustering step to aggregate pixel-wise predictions.

KPAM~\cite{gao2019kpam, manuelli2019kpam} presented a way to track category-level object keypoints. However, their system can only track single objects due to the integral pose regression step it relies on \cite{sun2018integral}. Furthermore, for the same reason, it can't deal with many keypoints of the same type.
KeyPose \cite{liu2020keypose} is an object keypoint detection method and dataset, which also uses stereo views of a scene. This method is only applicable in single object scenes; detected keypoints are not associated to objects, which makes tracking multiple objects infeasible. Further, KeyPose only works with objects that have unique keypoints. Modeling objects such as the valve in our experiments is not possible, as it has several ambiguous keypoints. This limitation is due to the spatial softmax operation that is used in the output heatmaps. The dataset collection method proposed by KeyPose relies on fiduciary tags that are placed in the scan environment. We propose and demonstrate the feasibility of a method that does not require modifying the environment and that relies solely on a calibrated robot with a camera.

%% file: 03-method.tex
\begin{figure*}[th]
    \centering
    \includegraphics[width=0.85\linewidth]{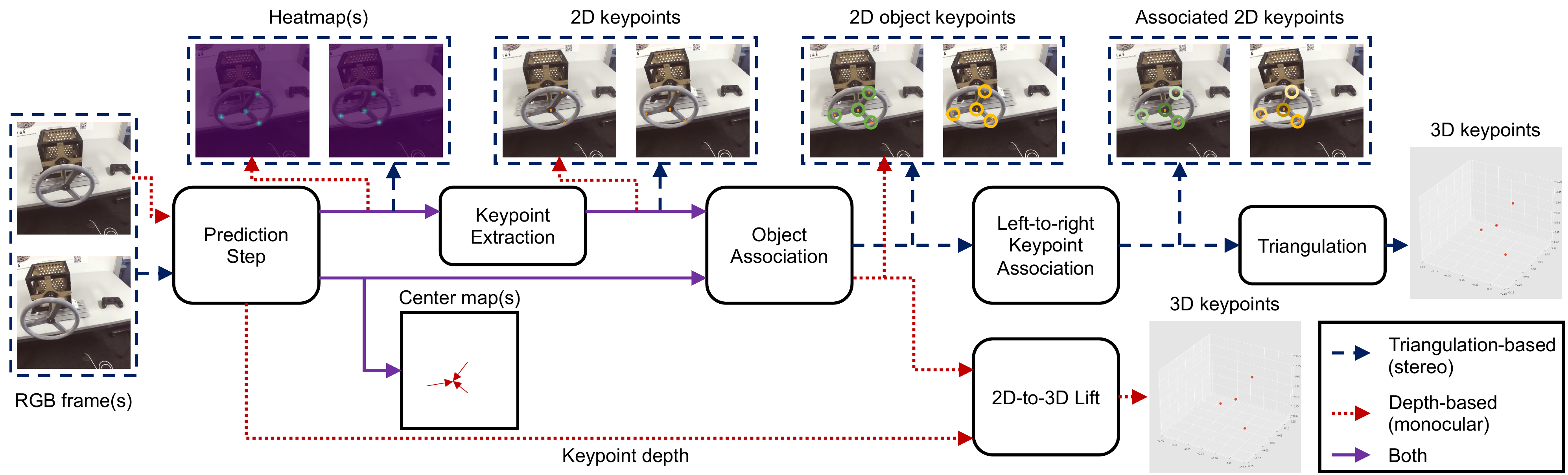}
    \caption{The components for both of the proposed keypoint tracking pipelines.}\vspace{-0.5cm}
    \label{fig:pipeline}
\end{figure*}

Here we describe the components of our framework: the hardware setup, the procedure to collect video sequences with camera poses, our algorithm to compute ground truth labels, a stereo multiple object keypoint detection pipeline and a variation that uses only a monocular RGB camera.

\subsection{Hardware Setup}

Our method requires a calibrated image sensor along with a way to control it into different known viewpoints. For this, we use a StereoLabs ZED Mini stereo RGB camera, mounted on the wrist of a Franka Emika Panda robot arm. The robot arm has accurate encoders at each joint which give readings on the position of each joint. Using a model of the robot and the position readings, we can accurately compute the position of the wrist, relative to the base frame of our robot.

We compute the intrinsic parameters and left-to-right-camera transformation of the stereo camera using Kalibr \cite{oth2013rolling}. The wrist-to-camera frame transformation we calibrate using the ethz-asl/hand\_eye\_calibration package \cite{Furrer2017FSR}.

\subsection{Data Collection}

For training a neural network to detect keypoints, we need a dataset of image frames and a set of keypoint locations for each image. To obtain these, we collect 30s long sequences while our robot scans the target objects, observing the objects from multiple viewpoints. We save the pose of both camera frames, relative to the base frame of the robot and the RGB frames from the left and right camera sensors. In the next section, we describe how we obtain the keypoints in image coordinates for each image.

\subsection{StereoLabel: Labeling and Generating a Dataset}
\label{sec:generate}
For each object category of interest, we define a set of keypoints that is most convenient for manipulating objects from the category. For example, in the case of coffee cups, we can define the keypoints to be the bottom center, center top and the outermost point on the handle of the cup (see Fig.~\ref{fig:stereo-label}). This allows us, if desired, to solve for the orientation of the cup. Or, we can grasp the cup by approaching the cup from the top center and grasping the side wall of the cup with a parallel jaw gripper. If there are several ambiguous keypoints, as is the case for the valve in our experiments (see Fig.~\ref{fig:valve}), then all occurrences of the keypoints are labeled and considered to be of the same type.

We developed a tool, StereoLabel, to label 3D keypoints from a sequence of images taken from different viewpoints. Fig.~\ref{fig:stereo-label}, shows the tool in use. The user is shown image frames from two viewpoints. The viewpoints are picked such that the z-axes of the image frames are as close to perpendicular as possible. The image frame is defined to be z-axis forward, y down and x to the right of the image. The user labels 2D keypoints on both frames by clicking on the keypoint location in the image. In case a specific keypoint is occluded or otherwise hard to pinpoint, the user can swap out either frame with a new one.

Once corresponding keypoints are labeled, we triangulate their 3D positions in the base frame of our robot using the homogeneous direct linear transformation method \cite{andrew2001multiple}. We backproject the triangulated points to both frames using each frame's projection matrix, so that the user can validate that the point was appropriately placed. The user can further validate correct placement by cycling through images in the sequence by pressing a button and checking the backprojected points. In addition to the labeled keypoints, we augment the set with one additional 3D keypoint; this is the average of all the other 3D keypoints which we call the center keypoint.

Once we have all the image sequences labeled and triangulated, we can generate a dataset for training a computer vision model. Fig.~\ref{fig:pipeline} shows the proposed triangulation-based (stereo) and depth-based (monocular) keypoint tracking pipelines. For each frame in a sequence, we create a set of ground truth heatmaps, one heatmap for each type of keypoint. Should there be several keypoints of the same type, we pack them onto the same heatmap. We compute the 2D image coordinate for each 3D keypoint by backprojection. We place a Gaussian distribution over the 2D keypoint location computed using an RBF kernel (output of the prediction step in Fig.~\ref{fig:pipeline}). Finally, we normalize each heatmap to have values in the range $[0, 1]$.

As there might be multiple objects in a frame, we compute 2D vector fields with vectors pointing from non-center 2D keypoints to the center keypoint. We compute one vector for output pixel having a non-zero heatmap value. With the center maps, we can associate keypoints to objects and detect multiple objects in a frame.

For the monocular version of our pipeline, we additionally compute a keypoint depth map containing the z-value of each 3D keypoint for each pixel within a fixed radius from each keypoint. 

\subsection{Learning the Keypoint Network}

We use a convolutional neural network (CNN) to predict the heatmaps, along with center maps. We use CornerNet-Lite \cite{law2019cornernet} as a backbone network. CornerNet-Lite is a stacked hourglass-style CNN architecture. The input is first downsampled through a series of convolutional layers and then upsampled through transposed convolutional layers in an hourglass module. Two hourglass modules are composed together.

We predict target maps with two prediction modules which take as input the output of each respective hourglass module. The prediction modules consist of three convolutional layers with batch normalization, 1$\times$1 kernels with stride 1 and relu activation functions, except for the last layer. We use sigmoid activation functions at the heatmap heads and no activation function for the center map and relu for the depth map.

The input to our network has size 511$\times$511 pixels and the output map resolution is 64$\times$64. We initialize the backbone network weights by pretraining on COCO~\cite{lin2014microsoft}.

We use three types of losses to train our network: a heatmap loss, a center loss and a depth loss. For the heatmap loss we use binary cross entropy:
\begin{align}
\mathbf{L}_h &= - \sum_{c=1}^C \sum_{i=1}^H \sum_{j=1}^W y_{cij} \log{p_{cij}} + (1 - y_{cij})\log{(1 - p_{cij})}.
\end{align}
$p_{cij}$ is the predicted heatmap value for a keypoint of type $c$ at output index $i, j$. $y_{cij}$ is the ground truth heatmap value for keypoint map $c$ at index $i, j$. $C$, $H$ and $W$ denote the amount of keypoint types, and the height and width of the output maps.

For the center loss, we simply use a smooth L1 loss:
\begin{align}
\mathbf{L}_c &= \sum_{c=1}^C \sum_{i=1}^H \sum_{j=1}^W \text{smooth\_L1}(\hat{\mathbf{c}}_{cij} - \mathbf{c}_{cij}) \check{y}_{cij},
\end{align}
where $\hat{\mathbf{c}}_{cij}$ is the center vector prediction for keypoint type $c$ at index $i, j$. $\mathbf{c}_{cij}$ is the corresponding ground truth center vector. $\check{y}_{cij}$ is a binary value denoting whether the heatmap value for keypoint type $c$ at index $i, j$ is nonzero. The smooth L1 loss is squared below a value of 1 and linear otherwise and is applied elementwise.

To enable using a monocular camera, we additionally need an estimate of how far along the z-axis each keypoint is. To do this, we predict a pixelwise depth estimate for each keypoint type. We learn this using an L1 loss function:
\begin{align}
    \mathbf{L}_d &= \sum_{c=1}^C \sum_{i=1}^H \sum_{j=1}^W \norm{z_{cij} - \hat{z}_{cij}}_1 \check{y}_{cij},
\end{align}
where $z_{cij}$ is the ground truth depth value for keypoint $c$ at location $i, j$, while $\hat{z}_{cij}$ is the corresponding estimate.

All losses are applied at both stages of the hourglass network and are combined by weighting parameters $\lambda_{\cdot}$:
\begin{align}
L = \lambda_h (L_{h1} + L_{h2}) + \lambda_c (L_{c1} + L_{c2}) + \lambda_d (L_{d1} + L_{d2}).
\end{align}
$L_{h1}$ denotes the heatmap loss at the first hourglass, $L_{h2}$ for the second hourglass, $L_{c1}$ the center loss for the first hourglass and so forth. When training the triangulation-based pipeline, we set the depth loss weight $\lambda_d$ to 0 to disregard it entirely. We train our network using the dataset generated in Section~\ref{sec:generate}.

\subsection{Keypoint Extraction}

At runtime, we extract keypoint locations from the heatmaps by first applying a version of non-maxima supression, where we zero the non-maximum values in 5$\times$5 regions surrounding each location. We then zero out all values below a threshold of $0.25$. From each of the remaining heatmap values, we compute keypoint locations by weighing image indices by the predicted heatmap density in a 5$\times$5 region on the unprocessed heatmap predictions centered at the maxima location.

For each non-center keypoint, we compute the object center estimate by summing the center vector with the corresponding image index. We associate each keypoint with the center keypoint closest to the keypoint's predicted center position in pixel coordinates.

\subsection{Keypoint Association and Triangulation}

After predicting and extracting keypoints in left and right image frames, we need to associate each keypoint in the left frame, to its counterpart in the right frame. To do this, we select the keypoint in the right image where $\boldsymbol{x'}\boldsymbol{F}\boldsymbol{x}$ is below a cutoff value of $32.0$. $\boldsymbol{F}$ is the fundamental matrix derived from the camera calibration, $\boldsymbol{x}$ is the homogeneous pixel coordinates of the keypoint in the left image and $\boldsymbol{x}'$ is the homogeneous pixel coordinates of the keypoint in the right image.

If several keypoints match, which happens when two keypoints are on the epipolar line, we shift the point by a fixed amount and pick the closest match. The fixed shift is equivalent to the difference in pixel coordinates between a point, projected onto both the left and right image frames, that is 60cm in front of the center of the left camera frame.

Finally, we triangulate the 3D location of the keypoints using the same direct linear transformation method used when creating the dataset.

\subsection{2D-to-3D}

In the monocular version of our pipeline, we use the depth prediction, combined with the camera matrix $\mathbf{K}$ to compute the 3D point $\mathbf{X}$ corresponding to the 2D detection $\mathbf{x}$:
\begin{align}
    \mathbf{X} = \mathbf{K}^{-1} \mathbf{x}  \hat{z},
\end{align}
where $\hat{z}$ is the depth estimate for keypoint $\mathbf{x}$.

%% file: 04-experiments.tex
We are interested in the following questions:
\begin{itemize}
    \item Can we use our method to quickly build up object keypoint tracking datasets?
    \item Can we train our keypoint tracking method on an amount of data that can be easily collected by one user?
    \item Is the object tracking performance good enough to enable robotic manipulation?
\end{itemize}

\subsection{Valve: Single Object Tracking}

\begin{figure}[t]
    \centering \vspace{-0.4cm}
    \hfill
    \begin{subfloat}[\label{fig:valve}]
        {\includegraphics[width=0.4\columnwidth]{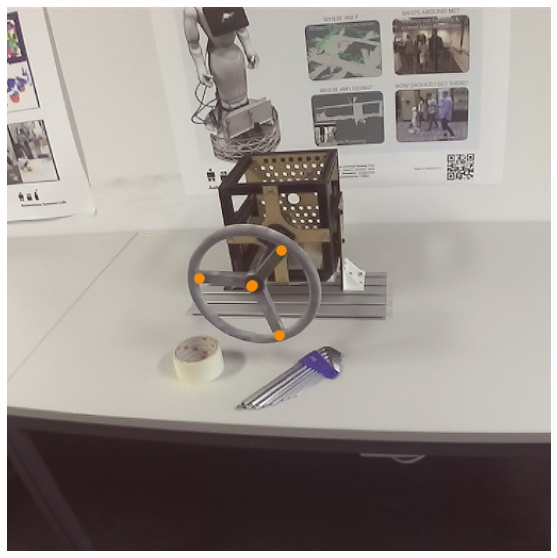}}
    \end{subfloat}
    \hfill
    \begin{subfloat}[\label{fig:cups}]
        {\includegraphics[width=0.4\columnwidth]{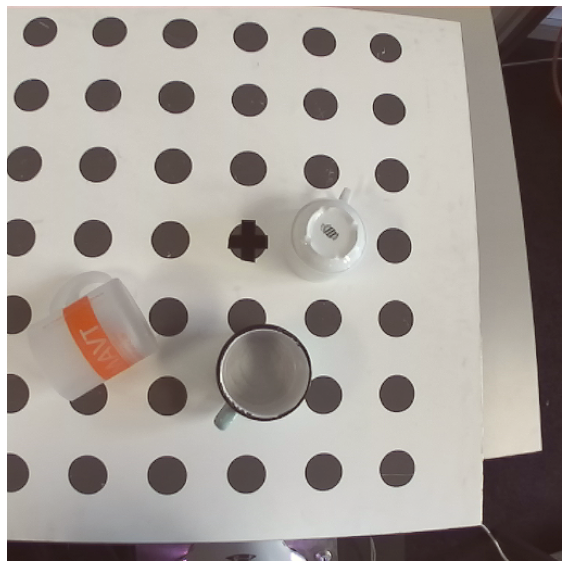}}
    \end{subfloat}
    \hfill
    \caption{(a) Valve setup showing keypoints for the value. (b) An image from the cup tracking scene.}\label{figure:categories}
\end{figure}

In this experiment, we track a valve with three spokes. We define four keypoints: one at the center hub of the valve, and three at the front center points where the spokes meet the rim of the valve, shown in Fig.~\ref{fig:valve}. The three keypoints at the rim are indistinguishable from each other, and are thus considered to be of the same type and packed onto the same heatmap.

We collect 50 sequences of 30s using our data collection method, which we label using StereoLabel. The sequences differ in object arrangement, clutter, occlusion, background and lighting conditions. We split the resulting dataset into 45 sequences for training and 5 sequences for testing. 
\subsection{Label Accuracy}

Our semi-automatic labeling approach has a few sources of error: synchronization between camera frames and joint encoder readings, intrinsics calibration error and hand-eye calibration error. These all result in some error in the triangulation and reprojection steps of our pipeline. Without the ground truth 3D keypoint locations, we instead manually label 2D keypoints frame-by-frame on 100 randomly sampled frames in our valve dataset and compare the human labels to ones produced by our system. This allows us to quantify how much the 2D keypoint labels drift as we observe the target object from different viewpoints. As the user does not always place the keypoints perfectly, even the human labels will have some error. We therefore establish a baseline by doing two manual frame-by-frame labeling passes to measure the variance.

\subsection{KeyPose}

We compare our method against KeyPose \cite{liu2020keypose}. KeyPose doesn't support multiple objects nor is it possible to detect keypoints on objects with multiple keypoints of the same type. We therefore can't run KeyPose on our datasets, but instead opt to evaluate our method on the KeyPose mugs dataset.

\subsection{Cups: Multi-object Tracking}

In this experiment, we seek to track up to four cups simultaneously in a scene. We collect a dataset with 100 sequences observing the cups from various viewpoints, varying the number of cups between 1 and 4 in the scene, changing the clutter, lighting conditions and background of the scene across sequences. For each scene, we randomly select between 1 and 4 cups from a set of 25 different cups. We split this dataset into 87 sequences for training and the rest for testing. We split the sequences such that 2 cups only ever occur in the test set.

%% file: 05-results.tex
\subsection{Valve: Single Object Tracking}

We timed how long it takes to label a sequence of images. Labeling a pair of valve images took us just under 15 seconds. With 50 sequences, this makes for a total of $\sim$12 minutes and 30 seconds to label all 19'507 frames in our dataset.

\begin{table}[b]
\caption{Results from the valve and cup tracking experiments.}
\begin{tabularx}{\linewidth}{llllll}
\toprule
\multicolumn{6}{c}{\textbf{Valve}} \\
\bottomrule
\textbf{Method} & \textbf{Mean} (cm) & \textbf{xy} (cm) & \textbf{$< 3$ cm} & \textbf{25th} (cm)& \textbf{75th} (cm)\\ \midrule
GT & 0.39 & 0.18 & 99.3\% & 0.17 & 0.30 \\
Stereo & 3.63 & 1.43 & 59.5\% & 1.29 & 4.26  \\
Mono & 2.99 & 1.06 & 65.0\% & 1.61 & 3.55\\
\toprule
\multicolumn{6}{c}{\textbf{Cups}} \\
\bottomrule
\textbf{Method} & \textbf{Mean} (cm)& \textbf{xy} (cm)& \textbf{$< 3$ cm} & \textbf{25th} (cm) & \textbf{75th} (cm) \\ \midrule
GT & 1.14 & 0.39 & 97.7\% & 0.09 & 0.22 \\
Stereo & 6.71 & 2.24 & 68.5\% & 1.29 & 3.53 \\
Mono & 3.1 & 1.56 & 62.2\% & 1.43 & 4.02 \\
\bottomrule
\end{tabularx}
\label{table:results}
\end{table}

Table \ref{table:results} shows the keypoint tracking performance on a held out test set. \textbf{Mean} refers to the mean error of the 3D keypoints in centimeters. \textbf{xy} is the mean error, disregarding the depth axis in the left camera frame of reference. \textbf{$< 3$ cm} is the percentage of measurements that were within 3 centimeters of the labeled ground truth location. \textbf{25th} and \textbf{75th} respectively denote the 25th and 75th percentiles of the combined keypoint errors. GT refers to the tracking performance using ground truth heatmaps and center maps as input with the stereo pipeline, Stereo is the stereo pipeline with a learned model, Mono is using only the left view of our stereo camera and our monocular pipeline.

Both the stereo- and depth-based pipelines perform reasonably well. For both pipelines, errors are within the range of the width of a parallel jaw gripper, and much smaller in scale than the size of the object.

\subsubsection{Valve Manipulation}

We deployed our keypoint tracking system on a mobile manipulation system. The goal of the experiment was to use the system to rotate the valve in Fig.~\ref{fig:valve} using the manipulator. In this case, we know the type of the valve and have a CAD model of it. We first detect the valve using keypoint tracking, and when we have detected all four keypoints (center and three spokes), we further refine the pose using ICP to match the depth readings from our camera to the object model. After refining the object pose, we command the arm to track a trajectory that rotates the valve. See the supplementary video for a successful completion of the task.

\subsection{Label Accuracy}
Comparing our generated keypoints to a manually labeled dataset, we found that the mean label difference is 6.3 pixels on average with a standard deviation of 3.4 on images with a size of 1280x720. Comparing two manually and separately labeled instances of the same datasets yield a mean difference of 2.9 pixels with a standard deviation of 1.7 pixels. While the manually labeled examples have slightly less variance, they are of the same order of magnitude for both methods.

\subsection{KeyPose}

\begin{table}[b]
\centering
\caption{Our monocular pipeline on the KeyPose mugs dataset.}
\begin{tabular}{lcc}
\toprule
\textbf{Method} & \textbf{MAE (cm)} & \textbf{$< 2$ cm} \\ \midrule
KeyPose & 1.6 & 78.6 \\
Ours monocular & 2.0 & 66.4 \\
Ours Stereo & 1.9 & 69.7 \\
\bottomrule
\end{tabular}
\label{table:keypose}
\end{table}

Table \ref{table:keypose} shows results on the KeyPose mugs dataset evaluating on the unseen $mugs\_0$ instance. Keypose performs slightly better. We attribute this to its more restricted problem formulation and the learned stereo image fusion employed in its network architecture.

\subsection{Cups: Multi-object Tracking}

It took us an average of 19 seconds to label a scene with 2 cups. Which makes for a total of roughly 32 minutes to label the 66'419 frames in our dataset.

Table~\ref{table:results} shows the accuracy when tracking multiple cups on a held out test set. Similarly to the valve tracking experiment, the error is larger in the depth direction. Performance of both the stereo and depth pipelines is acceptable, i.e. errors are within the width of a parallel jaw gripper. Performance is slightly worse than on the valve tracking experiment. However, we note that this is a harder task with several different objects and significantly more keypoint occlusion.

\begin{table}[b]
\centering
\caption{Time spent on each step of the stereo pipeline.}
\begin{tabular}{l c}
\toprule
\textbf{Stage}          & \textbf{Mean time (ms)}                \\ \midrule
\textbf{Prediction}     & 32.9 \\
\textbf{Keypoint extraction}   & 6.8\\
\textbf{Object association}   & 0.62 \\
\textbf{Left-to-right association} & 0.1 \\
\textbf{Triangulation} & 0.2 \\
\bottomrule
\end{tabular}
\label{timings}
\end{table}

Failure modes for both pipelines include misdetecting a keypoint or associating a keypoint with the wrong object. Failure modes of the stereo pipeline also include misassociating keypoints from left-to-right and a bad triangulation due to slightly misdetected 2D keypoints. Additionally, both approaches fail when two keypoints of the same type align, either from the same or different objects, and occlude each other. In such cases, the keypoints will get detected as one and the center prediction might point toward either object in the case of multiple objects.

Table~\ref{timings} shows how long each step of our stereo pipeline takes on average for our implementation. The measurements were made on a computer with an Nvidia RTX 2080 GPU and AMD EPYC 7742 CPU.

%% file: 06-conclusions.tex
In this paper, we presented a method to quickly collect and label object keypoint tracking datasets and a system that learns to recover the labels at runtime on unseen examples. We showed that we fully rely on calibration to avoid having to place markers in the environment. In experiments, we showed that we can generate accurate object keypoint labels much quicker than using a 2D labeling approach, while also annotating the z-dimension and without having to create segmentation maps. We showed that our presented system is able to detect keypoints on multiple objects simultaneously at real-time rates, using the produced datasets. We showed that it can be successfully used as part of a system to solve real world manipulation tasks. 

While we presented two different keypoint tracking algorithms, the actual keypoint and object detection algorithm can be replaced with any other pipeline, as long as the labels can be derived from our data collection method. Additional information about the objects could also be used to further improve the estimated keypoints. The Perspective-n-Point algorithm could be used for known objects or a category-level object model could be fitted to the keypoint detections to further improve them, similar to what is used in \cite{yang2021dynamical}. Additional computation could be traded for accuracy by predicting keypoint heatmaps at a higher resolution. The 64x64 pixel output resolution we used is quite limiting.

One weakness of our data collection method, is that it relies on measurement timestamps from different sensors. On our platform, these are not hardware synchronized. 
Some cameras can be tightly synchronized, while triggering boards such as the VersaVIS board exist \cite{tschopp2020versavis} which are able to synchronize several cameras and IMUs. Extending these to cover other types of sensors, such as joint encoders, would improve the usability and accuracy of our proposed method. 

Finally, when collecting our data, we manually guide the robot into different viewpoints. This could be automated. Furthermore, the robot could semi-autonomously improve upon an initial learned model using an active learning type approach. Different models and viewpoints could be used to bootstrap a dataset, similar to what is done in \cite{simon2017hand}.